\documentclass[10pt,twocolumn,letterpaper]{article}

\usepackage{cvpr}              

\usepackage[dvipsnames]{xcolor}

\definecolor{cvprblue}{rgb}{0.21,0.49,0.74}
\usepackage[pagebackref,breaklinks,colorlinks,citecolor=cvprblue]{hyperref}

\usepackage[accsupp]{axessibility}

\usepackage{algorithm}
\usepackage{algorithmic}
\usepackage{subcaption}
\usepackage{caption,subcaption}
\usepackage{amsmath}
\usepackage{amssymb}
\usepackage{mathtools}
\usepackage{amsthm}
\usepackage{multirow}
\usepackage{algorithm,algorithmic}

\usepackage{pifont}
\usepackage{booktabs}
\usepackage{mathtools}
\usepackage{amsthm}
\usepackage{multirow}

\usepackage{pifont}
\usepackage{wrapfig}
\usepackage{lipsum}
\newcommand\blfootnote[1]{
    \begingroup
    \renewcommand\thefootnote{}\footnote{#1}
    \addtocounter{footnote}{-1}
    \endgroup
}

\newtheorem{remark}{Remark}

\def\paperID{14629} 
\def\confName{CVPR}
\def\confYear{2024}

\title{Enhancing Multimodal Cooperation via Sample-level Modality Valuation}

\author{
\textbf{Yake Wei}\textsuperscript{1}, 
\textbf{Ruoxuan Feng}\textsuperscript{1}, 
\textbf{Zihe Wang}\textsuperscript{1,2}, 
\textbf{Di Hu}\textsuperscript{1,2,*}
\vspace{0.5em}
\\
\textsuperscript{1}Gaoling School of Artificial Intelligence, Renmin University of China, Beijing\\
\textsuperscript{2}Beijing Key Laboratory of Big Data Management and Analysis Methods, Beijing\\
\{yakewei, fengruoxuan, wang.zihe, dihu\}@ruc.edu.cn
}

\begin{document}

\maketitle

\begin{abstract}
One primary topic of multimodal learning is to jointly incorporate heterogeneous information from different modalities. However, most models often suffer from unsatisfactory multimodal cooperation, which cannot jointly utilize all modalities well. Some methods are proposed to identify and enhance the worse learnt modality, but they are often hard to provide the fine-grained observation of multimodal cooperation at sample-level with theoretical support. Hence, it is essential to reasonably observe and improve the fine-grained cooperation between modalities, especially when facing realistic scenarios where the modality discrepancy could vary across different samples. To this end, we introduce a sample-level modality valuation metric to evaluate the contribution of each modality for each sample. Via modality valuation, we observe that modality discrepancy indeed could be different at sample-level, beyond the global contribution discrepancy at dataset-level. We further analyze this issue and improve cooperation between modalities at sample-level by enhancing the discriminative ability of low-contributing modalities in a targeted manner. Overall, our methods reasonably observe the fine-grained uni-modal contribution and achieve considerable improvement. The source code and dataset are available at \url{https://github.com/GeWu-Lab/Valuate-and-Enhance-Multimodal-Cooperation}.\blfootnote{\noindent
\textsuperscript{*}Corresponding author.
}
\end{abstract}

\section{Introduction}
Humans are surrounded by messages of multiple senses, including vision, auditory and tactile, bringing us a comprehensive perception. Inspired by this multi-sensory integration phenomenon, learning from multimodal data has raised attention in recent years. Recent researchers have well summarized the wide range of applications of multimodal learning and looked at its future development~\cite{wei2022learning}. One primary topic in multimodal learning is how to jointly incorporate multiple heterogeneous information. In the early, researchers tried to achieve the union of multiple modalities via different perspectives, like probabilistic theory based dynamic Bayesian networks~\cite{nefian2002dynamic} and multimodal Restricted Boltzmann Machines inspired by thermodynamic~\cite{ngiam2011multimodal}

\begin{figure}[t]
\centering
\centering
\includegraphics[width=0.4\textwidth]{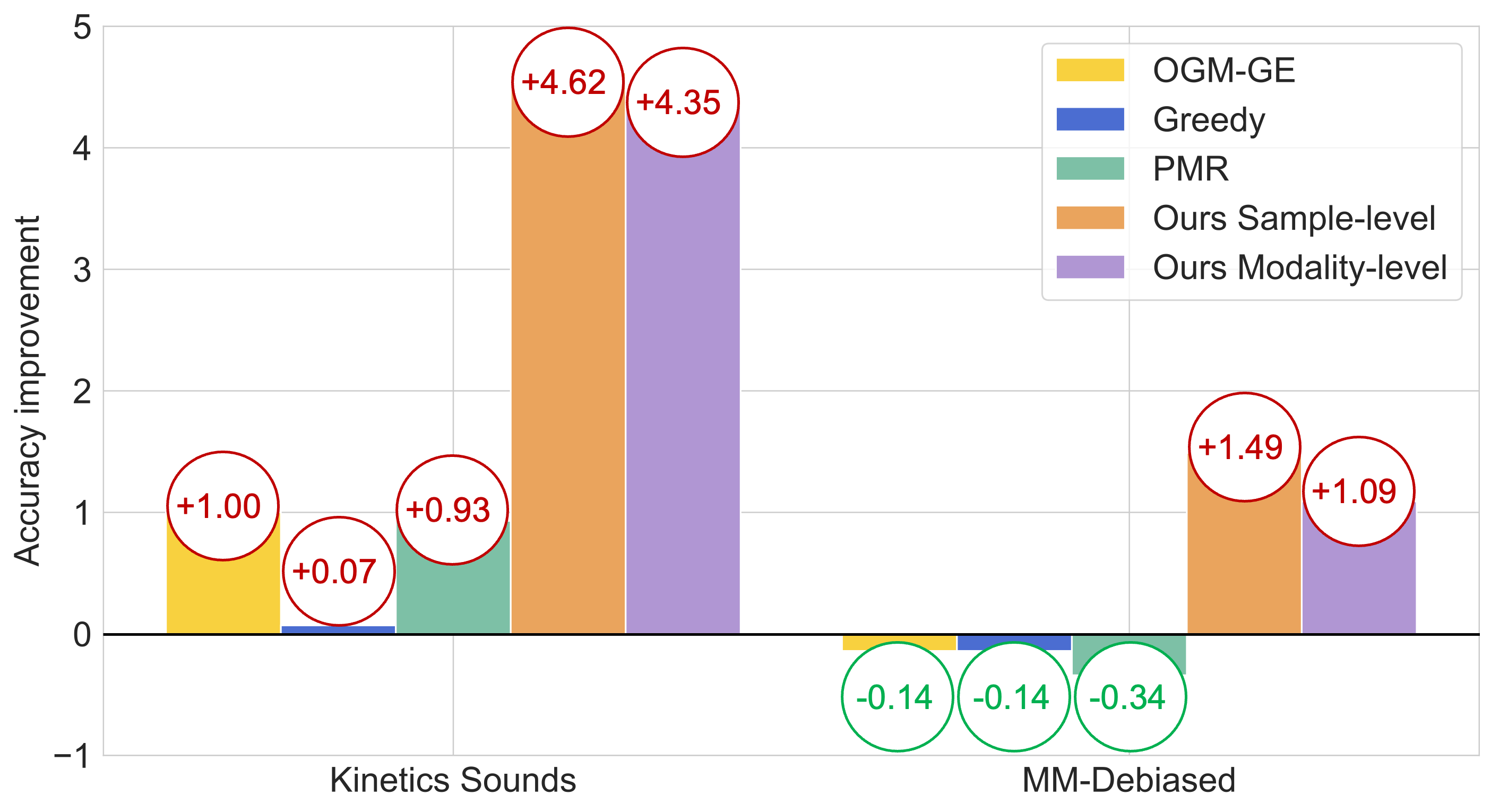}
\vspace{-0.5em}
\caption{Accuracy improvement compared with joint training baseline of imbalanced multimodal learning methods, on Kinetics Sounds and our proposed MM-Debiased dataset. Other methods: OGM-GE~\cite{peng2022balanced}, Greedy~\cite{wu2022characterizing} and PMR~\cite{fan2023pmr}. }
\vspace{-1.5em}
\label{fig:teaser-bias}
\end{figure}

\begin{figure*}[t]
\centering
    \begin{subfigure}[t]{.22\textwidth}
			\centering
			\includegraphics[width=\textwidth]{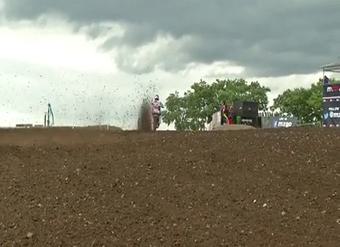}
			\caption{Visual of \emph{S.1} of \emph{motorcycling}.}
			\label{fig:teaser_audio}
	\end{subfigure}
    \begin{subfigure}[t]{.22\textwidth}
			\centering
			\includegraphics[width=\textwidth]{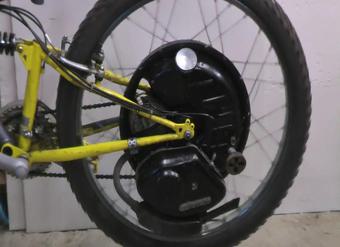}
			\caption{Visual of \emph{S.2} of \emph{motorcycling}.}
			\label{fig:teaser_visual}
	\end{subfigure}
	\begin{subfigure}[t]{.22\textwidth}
			\centering
			\includegraphics[width=\textwidth]{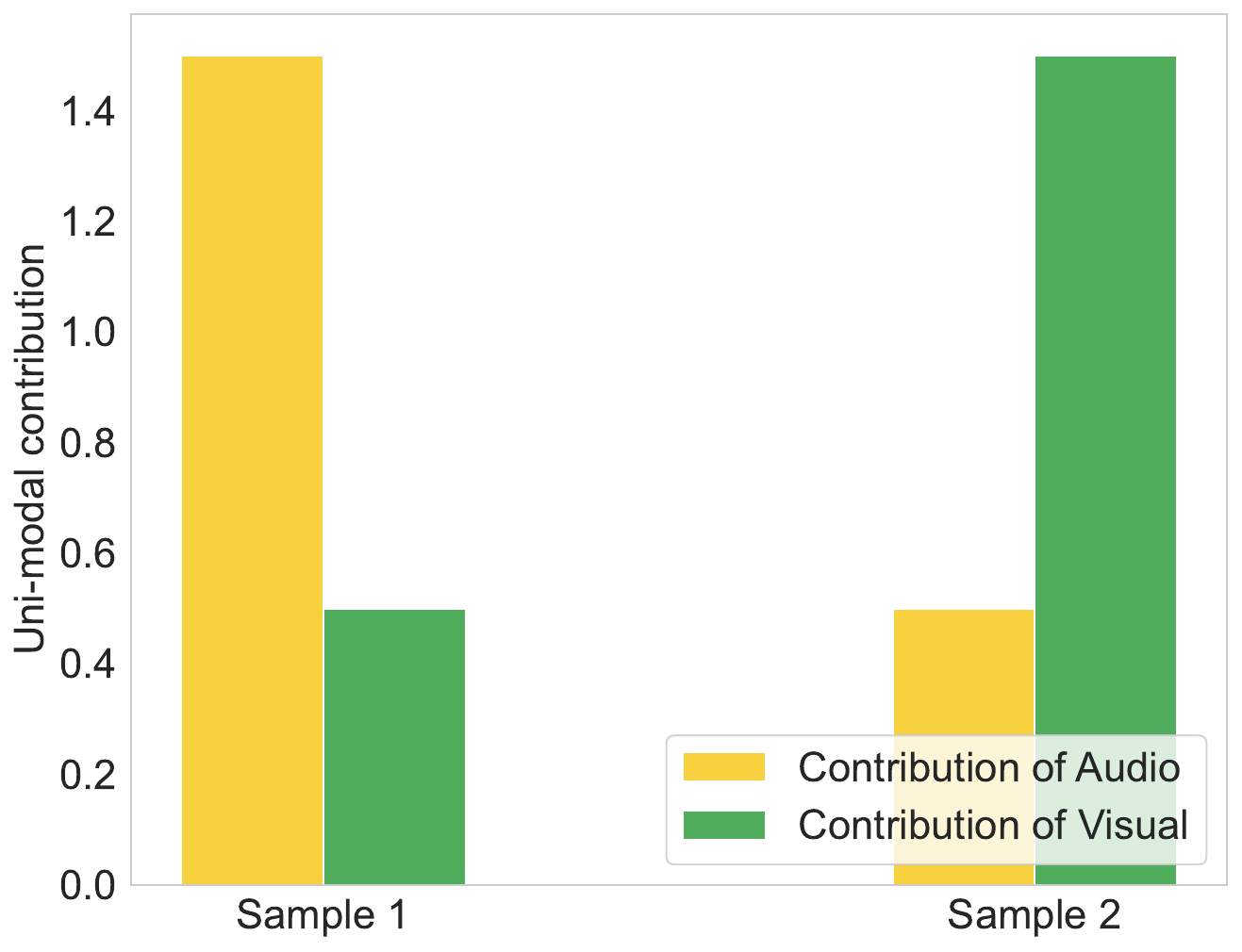}
			\caption{Valuation of \emph{S.1} and \emph{S.2}.}
			\label{fig:teaser-sample}
	\end{subfigure}
         \begin{subfigure}[t]{.22\textwidth}
			\centering
			\includegraphics[width=\textwidth]{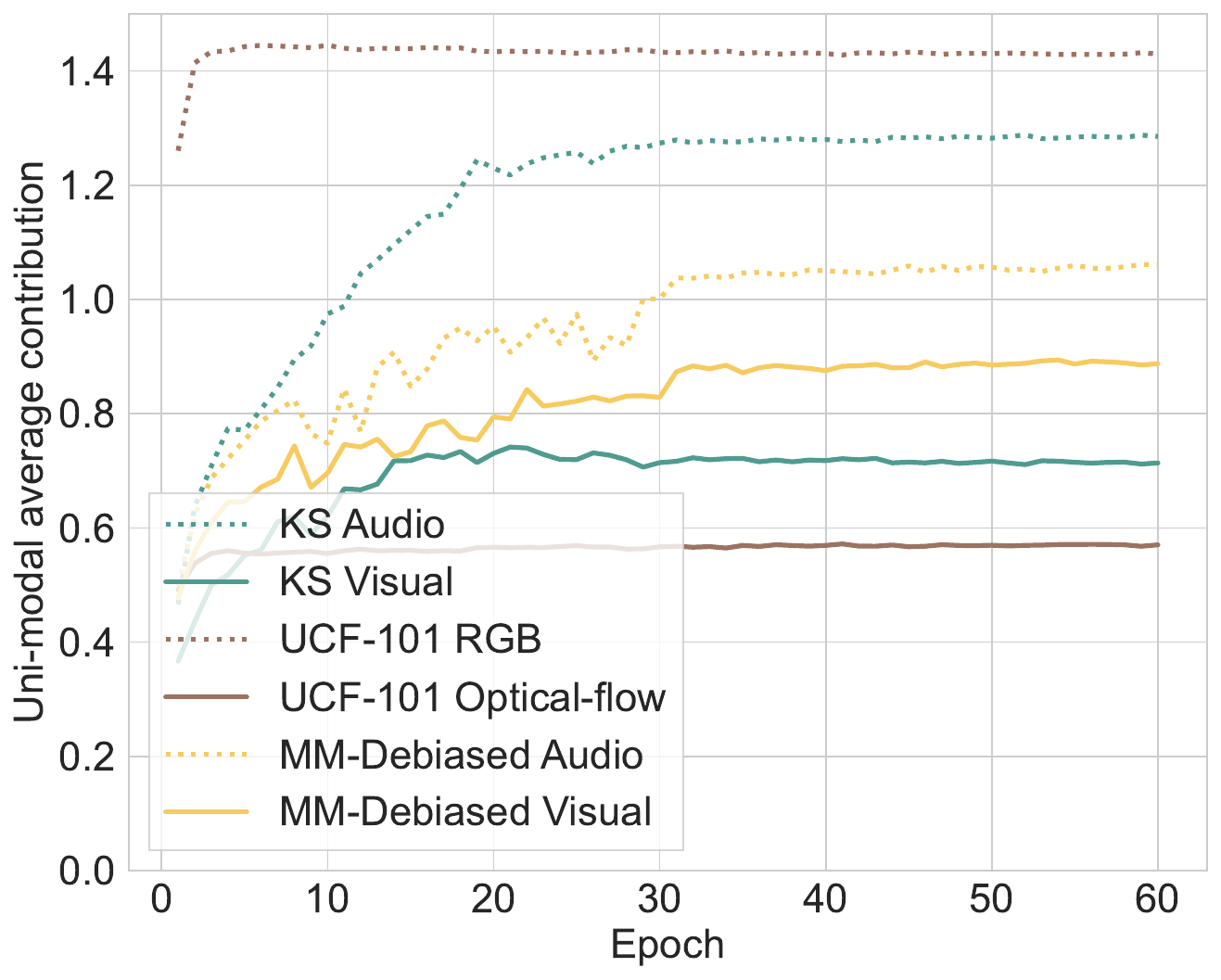}
			\caption{Avg. Contribution of dataset.}
			\label{fig:shap-dataset}
	\end{subfigure}
 \vspace{-0.8em}
    \caption{\textbf{(a-b):} Audio-visual samples of \emph{motorcycling} category. \textbf{(c):} Our modality valuation of \emph{S.1} and \emph{S.2}. \emph{S.1} and \emph{S.2} denotes \emph{Sample 1} and \emph{Sample 2} respectively. \textbf{(d):} Uni-modal average contribution over all training samples of different dataset. Our proposed MM-Debiased dataset has less global discrepancy at dataset-level, compared with other curated dataset.}
    \vspace{-1.2em}
    \label{fig:teaser-two}
\end{figure*}

As deep learning improves by leaps and bounds, deep neural networks with the capacity to learn representation from a large amount of data have been used extensively in multimodal learning~\cite{wang2021survey}. 
Although the deep-based methods have revealed effectiveness, recent studies have found the imbalanced multimodal learning problem where most existing models often have unsatisfactory multimodal cooperation, which cannot jointly utilize all modalities well~\cite{huang2022modality,wu2022characterizing}. But deep models' lack of interpretability makes it hard to observe what role each modality plays in the final prediction, and then accordingly adjust the uni-modal training. Some methods have been proposed to identify and improve the training of worse learnt modality with the help of output logits or the scale of gradient~\cite{peng2022balanced,wu2022characterizing}. These empirical strategies only consider the global modality discrepancy at dataset-level, and achieve improvement on the common curated dataset (as Kinetics Sounds dataset in Figure~\ref{fig:teaser-bias}). \emph{However, under realistic scenarios, the modality discrepancy could vary across different samples.} 
For example, Figure~\ref{fig:teaser_audio} and~\ref{fig:teaser_visual} show two audio-visual samples of \emph{motorcycling} category. The motorcycle in \emph{Sample 1} is hard to observe while the wheel of motorcycle in \emph{Sample 2} is quite clear.
This could make audio or visual modality contribute more to the final prediction respectively for these two samples. This fine-grained modality discrepancy is hard to perceive by existing methods. Hence, how to \emph{reasonably} observe and improve multimodal cooperation at sample-level is still expected to be resolved.

To this end, we introduce a sample-level modality valuation metric, to observe the contribution of each modality during prediction for each sample. The Shapley value from game theory~\cite{Shapley195317AV}, which aims to fairly distribute the benefits based on the contribution of each player, provides the theoretical support of our valuation. 
Via valuating uni-modal contribution, we observe that the experiment results unsatisfactorily fail to meet the expectation that each modality has its irreplaceable contribution. Firstly, as Figure~\ref{fig:shap-dataset}, for existing curated dataset including Kinetics Sounds and UCF-101, we verify that the contribution of one modality tends to overwhelm others globally at the dataset-level. More importantly, as Figure~\ref{fig:teaser-sample}, with our sample-level modality valuation, we observe that modality discrepancy indeed could be different across samples, beyond the global contribution discrepancy at dataset-level. To highlight this sample-level modality discrepancy, we propose the global balanced MM-Debiased dataset where the dataset-level modality discrepancy is no longer significant (as Figure~\ref{fig:shap-dataset}). Not surprisingly, existing imbalanced multimodal learning methods which only consider dataset-level discrepancy \emph{fail} on MM-Debiased dataset, as shown in Figure~\ref{fig:teaser-bias}.

Based on the above empirical results, we first analyze the effect of the modality with clearly lower contribution in a sample and find that \emph{its presence would potentially increase the risk that the multimodal model collapses to one specific modality}. Hence, it is urgent to recover the suppressed contribution of these low-contributing modalities. To alleviate the above problem, we further analyze the correlation between uni-modal discriminative ability and its contribution, then find that \emph{enhancing the discriminative ability of low-contributing modality during training could indirectly improve its contribution in a sample, and accordingly enhance multimodal cooperation}. Therefore, we propose to train the low-contributing modality in a sample in a targeted manner based on the contribution discrepancy between modalities. Specifically, we first valuate the uni-modal contribution at sample-level via our Shapley-based modality valuation metric. Then the input of identified low-contributing modalities is re-sampled with a dynamical frequency, determined by the exact contribution discrepancy, to improve its discriminative ability targetedly. Considering the computational cost of sample-wise modality valuation, we also propose the more efficient modality-level method.

As Figure~\ref{fig:teaser-bias}, our methods considering the sample-level modality discrepancy achieve considerable improvement on both existing curated and our global balanced dataset,
Overall, our contributions are as follows. \textbf{Firstly,} we introduce a sample-level modality valuation metric and further analyze the low-contributing modality issue, which could worsen the multimodal cooperation.
\textbf{Secondly}, methods are proposed to strengthen low-contributing modalities, reasonably improving multimodal cooperation. \textbf{Thirdly}, we propose the MM-Debiased dataset with fine-grained multimodal discrepancy, which is closer to the realistic scenario.

\section{Related works}

\noindent \textbf{Imbalanced multimodal learning.}
Recent studies have found the multimodal model has a preference for specific modality~\cite{huang2022modality}. Several methods have been proposed to ease this problem by improving the optimization of worse learnt modality~\cite{peng2022balanced,wu2022characterizing,fan2023pmr,xu2023mmcosine,yang2023quantifying}. These methods often control uni-modal optimization by estimating the discrepancy of the training stage or performance between modalities. However, their estimation could be hard to observe modality discrepancy at sample-level, handling realistic scenarios where the performance difference of each modality could vary across samples.
In this paper, we go a step further, reasonably valuating the uni-modal contribution at sample-level using the introduced Shapley-based metric. This fine-grained modality valuation metric could guide us to solve the imbalanced multimodal learning problem better.

\noindent \textbf{Game theory in machine learning.}
Researchers have adapted the game theory to formulate and solve machine learning problems~\cite{freund1996game,wang2017generative,gemp2020eigengame}. For example, game theory has been used to explain the algorithm effectiveness of AdaBoost ~\cite{freund1996game}. 
Similarly to us, Hu et al.~\cite{hu2022shape} use the Shapley value to evaluate the overall contribution of individual modalities for the whole dataset.
But they cannot capture the modality contribution at sample-level and do not provide further analysis or methods. 
In this paper, we not only introduce sample-level modality valuation, but also further analyze and alleviate the low-contributing modality issue.

\section{Method}

\subsection{Model formulation}
\label{sec:formulation}
In this paper, we consider the multimodal discriminative task. Concretely, each sample $x=(x^1,x^2,\cdots,x^n)$ is with $n$ modalities. And $y$ is the ground truth label of sample $x$. For simplicity, the input of modality $i$ of specific sample $x$ is denoted as $x^i$. $N = \{x^1,x^2,\cdots,x^n\} $ is a finite, non-empty set of all modalities. Denote the multimodal model as $H(\cdot)$. Suppose $C$ is the set of input modalities for the model, where $C \subseteq N$. When taking modalities in $C$ as the input, the final prediction is $\widehat{y_C}=H(\cup x^i, x^i  \in C)$. It should be noted that we have no assumption of multimodal fusion design, therefore the following modality valuation is not limited to simple fusion strategy.

\subsection{Fine-grained modality valuation}
\label{sec:valuation}
In multimodal learning, each modality is expected to fully demonstrate its irreplaceable contribution, since different modalities are considered with complementary information. Based on realistic scenarios, the modality contribution discrepancy could vary across different samples. 
Hence, it is necessary to valuate the uni-modal contribution in the multimodal model at sample-level, and accordingly improve multimodal cooperation. In this paper, we introduce a Shapely-based metric fine-grained modality valuation metric, to observe the uni-modal contribution for the multimodal prediction at sample-level. 

Concretely, for each sample $x$, we first have $v$ as a function to map the multimodal prediction to its benefits:
\begin{equation}
\label{equ:v-function}
v(C)=\begin{cases}
 |C|& \text{ if } \widehat{y_C} = y,\\ 
 0& \text{otherwise}.
\end{cases}
\end{equation}
When predicting correctly, the benefits of multimodal prediction with input $C$ is the number of input modalities. 

After formulating the prediction benefits, to consider the contribution of each modality under all cases, let $\Pi_{N}$ denote the set of all permutations of $N$. Since the number of modalities is $n$, there is $|\Pi_{N}|=n!$. For modality $i$ of sample $x$, given a permutation $\pi \in \Pi_{N}$, we denote by $S_{\pi}(x^i)$ the set of all predecessors of it in $\pi$, \emph{i.e.,} we set $S_{\pi}(x^i)= \{x^j \in N | \pi(x^j)<\pi(x^i)\}$. The \textit{marginal contribution} of modality $i$ of sample $x$ with respect to a permutation $\pi$ is denoted by $\Delta_{\pi}(x^i)$ and is given by:
\begin{equation}
\Delta_{\pi}(x^i)=v(S_{\pi}(x^i) \cup {x^i})  - v(S_{\pi}(x^i)).
\end{equation}
This quantity measures how much modality $i$ increases the benefits of its predecessors in $\pi$ when it joins the permutation $\pi$. Since different combinations of modalities could have different results, we should consider all possible permutations to fully evaluate uni-modal contribution. Then, given $\Pi_{N}$ with $n!$ permutations, the \textit{final contribution} of modality $i$ is denoted by $\phi^i$ and is given by:
\begin{equation}
\label{equ:shap}
\phi^i=\frac{1}{n!}\sum_{\pi \in \Pi_{N}}\Delta_{\pi}(x^i).
\end{equation}
It should be noted that when considering all possible permutations, the sum of all uni-modal contribution $\sum^n_{i=1}\phi^i$ is in fact the benefits of multimodal prediction with all modalities as the input. Hence, for the normal model with all modalities as the input, when the contribution of one modality increases, the contribution of other modalities would accordingly decrease.
With this sample-level modality valuation metric, we could reasonably observe the uni-modal contribution for each sample.

\subsection{Low-contributing modality phenomenon}
\label{sec:low-contributing}
As Figure~\ref{fig:teaser-two}, both at sample-level and dataset-level, contribution of one modality could highly overwhelm others. In other words, the decision of multimodal model is dominated by one modality, remaining others low-contributing.

Here we analyze the effect of low-contributing modality to the benefits of normal multimodal model for sample $x$. Suppose the marginal contribution of modality in multimodal learning is non-negative, since the introduction of additional modality tends to not bring negative effects in practice. Based on the definition of uni-modal contribution for modality $i$, we have:
\begin{gather}
\phi^i=\frac{1}{n!}\sum_{\pi \in \Pi_{N}}\Delta_{\pi}(x^i),\\
\phi^i=\frac{1}{n!}\sum_{\pi \in \Pi_{N}}(v(S_{\pi}(x^i) \cup {x^i})  - v(S_{\pi}(x^i))), \\
n!\cdot \phi^i \geq \underbrace{(n-1)!\cdot (v(N)-v(N\backslash x^i))}_{\text{only consider cases $x^i$ is the last one, which have $(n-1)!$ permutations}}, \\
n\cdot \phi^i \geq  v(N)-v(N\backslash x^i),
\end{gather}
\begin{equation}
\label{equ:upper}
v(N)-v(N\backslash x^i) \leq n \cdot \phi^i.
\end{equation}

In addition, based on Equation~\ref{equ:v-function}, when predicting correctly, $v(N)=n$. We know that the minimum of $v(N \backslash x^i)$ is $0$. Then we have:
\begin{equation}
v(N)-v(N\backslash x^i) \leq n.
\end{equation}
However, based on Equation~\ref{equ:upper}, when $\phi^i < 1$, the upper bound of the difference between $v(N)$ and $v(N \backslash x^i)$ shrinks (\emph{i.e.,} $n \cdot \phi^i<n$). In other words, when the contribution of modality $i$, $\phi^i < 1$, the difference between $v(N)$ and $v(N\backslash x^i)$ decreases. The benefits of taking all modalities $N$ as the input becomes close to its subset $N \backslash x^i$. Assuming an extreme case where the contribution of all but one modality is very small, multimodal learning is close to uni-modal learning. Hence, it is essential to enhance the contribution of low-contributing modalities for each sample, improving multimodal cooperation.

\begin{remark} 
\label{coro:degradation}
Suppose the marginal contribution of modality is non-negative. For the normal multimodal model with all modalities of sample $x$ as the input, with benefits $v(N)=n$, when modality $i$ is low-contributing, \emph{i.e.,} $\phi^i<1$, the difference between $v(N)$ and $v(N\backslash x^i)$ decreases.
\end{remark}

In Remark~\ref{coro:degradation}, we suppose the marginal contribution of modality is non-negative. In practice, the introduction of additional modalities has been validated its benefit (non-negative effect) across different application tasks~\cite{liang2022foundations}. It also theoretically proves that multimodal learning provably performs better than uni-modal~\cite{huang2021makes}. These evidences indicate that the introduction of another related modality could not bring a negative impact in most cases. Based on this, we assume that the marginal contribution is non-negative.

To alleviate the above problem, we further analyze the correlation between uni-modal discriminative ability and its contribution and have Remark~\ref{coro:enhance}. Based on the analysis, strengthening the discriminative ability of low-contributing modality can improve its contribution to multimodal prediction. Correspondingly, the risk that multimodal model collapses to one specific modality is lowered\footnote{The specific theoretical analysis process of Remark~\ref{coro:enhance} is provided in the \emph{Supp. Materials}.}.

\begin{remark} 
\label{coro:enhance}
Suppose the marginal contribution of modality is non-negative and the numerical benefits of one modality's marginal contribution follow the discrete uniform distribution. Enhancing the discriminative ability of low-contributing modality $i$ can increase its contribution $\phi^i$. 
\end{remark}

\subsection{Re-sample enhancement strategy}
\label{sec:resample}
Based on Remark~\ref{coro:enhance}, enhancing the discriminative ability of low-contributing modality can expand its contribution. Hence, we propose to improve the discriminative ability of low-contributing modality during training by re-sampling its input in a targeted manner. 


Concretely, to ensure the basic discriminative ability, we first warm up the multimodal model for several epochs. Then, at each epoch, modality valuation is conducted once to observe uni-modal contribution for each sample. Subsequently, learning of the low-contributing modality can be targetedly improved via solely re-sampling its input. Here, we provide the fine-grained as well as effective sample-level re-sample method and the coarse but efficient modality-level re-sample method.

\begin{algorithm}
\caption{Sample-level method}
\label{alg:sample}
\begin{algorithmic}
\REQUIRE Original training dataset $\mathcal{D}$, training dataset with re-sample $\mathcal{D}^{\text{rs}}$, number of modalities $n$, model parameters $\theta$, training epoch $T$, warm-up epoch $F$.
\FOR{$t=0,\cdots,T-1$} 
    \IF{$t < F$}
        \STATE Update model parameters $\theta$ with dataset $\mathcal{D}$;
    \ELSE
        \STATE Initialize $\mathcal{D}^{\text{rs}}$: $\mathcal{D}^{\text{rs}}=\mathcal{D}$;
        \FOR{ each sample $x$ in  $\mathcal{D}$}
            \STATE Obtain uni-modal contribution $\{\phi^1,\phi^2,\cdots,\phi^n\}$ with Equation~\ref{equ:shap};
            \STATE Identify modality $i$ where $\phi^i<1$;
            \STATE Get frequency $s(x^{i})$ with Equation~\ref{equ:sample};
            \STATE Add $x^i$ with frequency $s(x^{i})$ into $\mathcal{D}^{\text{rs}}$;
        \ENDFOR
        \STATE Update model parameters $\theta$ with dataset $\mathcal{D}^{\text{rs}}$; 
    \ENDIF
\ENDFOR
\end{algorithmic}
\end{algorithm}
\vspace{-2em}

\subsubsection{Sample-level method}
After the modality valuation, the low-contributing modality $i$, $\phi^i<1$, for each sample, can be well distinguished and we can finely improve its learning at sample-level. Then the specific re-sample frequency is dynamically determined by the exact value of $\phi^i$ during training. Specifically, the re-sample frequency of modality $i$ for specific sample $x$ is:
\begin{equation}
\label{equ:sample}
s(x^{i})=\left\{\begin{array}{cl}
f_s (1-\phi^i) & \text { }\phi^i<1, \\
0 & \text { others, }
\end{array}\right.
\end{equation}
where $f_s(\cdot)$ is a monotonically increasing function. Utilizing this sample-level re-sample strategy, the low-contributing modality $i$ in sample $x$ is re-trained with a re-sample frequency inversely proportional to its contribution, \emph{i.e.,} the less the contribution is, the larger the re-sample frequency is. It is worth noting that only the low-contributing modality is taken during re-sampling, and inputs of other modalities are masked by $0$, to ensure targeted learning.

\subsubsection{Modality-level method}
Although sample-level modality valuation could provide fine-grained uni-modal contribution, there could be a high additional computational cost when the scale of dataset is quite large. Therefore, the more efficient modality-level method is proposed to lower cost. As Figure~\ref{fig:shap-dataset}, the low-contributing phenomenon has a dataset-level preference. For example, the average contribution of RGB modality over all training samples is obviously more than that of optical flow modality on the UCF-101 dataset. Hence, we propose a more coarse modality-level re-sample strategy, which estimates the average uni-modal contribution via only conducting modality valuation on the subset of training samples to reduce additional computational cost.

Concretely, we randomly split a subset with $Z$ samples in the training set to approximately estimate the average uni-modal contribution. Hence, the overall low-contributing modality $i$ with less average $\phi^i$, \emph{i.e.,} $\frac{\sum^Z_{k=1} \phi_k^i}{Z}$, can be identified. Then, other modalities remain unchanged, and modality $i$ in sample $x$ is dynamically re-sampled with specific probability $p(i)$ during training:
\begin{equation}
\label{equ:modality}
p(i)=f_m (\text{Norm}(d)),
\end{equation}
where $d=\frac{1}{n-1} (\sum^n_{j=1,j\neq i} (
\frac{\sum^Z_{k=1} \phi_k^j}{Z} -\frac{\sum^Z_{k=1} \phi_k^i}{Z}
))$. 
The discrepancy in average contribution between overall low-contributing modality $i$ compared to other modalities (\emph{i.e.,} $d$) is first $0-1$ normalized, then fed into $f_m(\cdot)$, a monotonically increasing function with a value between $0$ and $1$. This re-sample probability for overall low-contributing modality is proportional to the discrepancy in its average contribution compared to others. Compared to sample-level strategy, modality-level one is more efficient.

\begin{algorithm}
\caption{Modality-level method}
\label{alg:modality}
\begin{algorithmic}
\REQUIRE Original training dataset $\mathcal{D}$, training dataset with re-sample $\mathcal{D}^{\text{rs}}$, subset of training dataset $Z$, number of modalities $n$, model parameters $\theta$, training epoch $T$, warm-up epoch $F$.
\FOR{$t=0,\cdots,T-1$} 
    \IF{$t < F$}
        \STATE Update model parameters $\theta$ with dataset $\mathcal{D}$;
    \ELSE
        \STATE Initialize $\mathcal{D}^{\text{rs}}$: $\mathcal{D}^{\text{rs}}=\mathcal{D}$;
        \FOR{ each sample $x$ in  $Z$}
            \STATE Obtain uni-modal contribution $\{\phi^1,\phi^2,\cdots,\phi^n\}$ with Equation~\ref{equ:shap};
        \ENDFOR
        \STATE Identify overall low-contributing modality $i$;
        \STATE Get re-sample probability $p(i)$ with Equation~\ref{equ:modality};
        \FOR{ each sample $x$ in  $\mathcal{D}$}
            \STATE Add $x^i$ with probability $p(i)$ into $\mathcal{D}^{\text{rs}}$;
        \ENDFOR
        \STATE Update model parameters $\theta$ with dataset $\mathcal{D}^{\text{rs}}$; 
    \ENDIF
\ENDFOR
\end{algorithmic}
\end{algorithm}
\vspace{-1em}

\section{Experiment}
\subsection{Dataset and experimental settings}
\noindent \textbf{Kinetic Sounds} (KS)~\cite{arandjelovic2017look} is an action recognition dataset with two modalities, audio and video. This dataset contains 31 human action classes, which are selected from Kinetics dataset~\cite{kay2017kinetics}. It contains 19k 10-second video clips.

\noindent \textbf{UCF-101}~\cite{soomro2012ucf101} is an action recognition dataset with two modalities, RGB and optical flow. This dataset contains 101 categories of human actions. The entire dataset is divided into a 9,537-sample training set and a 3,783-sample test set according to the original setting.

\noindent \textbf{MM-Debiased} is an audio-visual dataset proposed by us, where dataset-level modality contribution discrepancy is not obvious. It covers 10 classes, and contains 11,368 training samples and 1,472 testing samples. Details about the dataset construction are provided in \emph{Supp. Materials}.


\noindent \textbf{Experimental settings.} When not specified, ResNet-18~\cite{he2016deep} is used as the backbone in experiments. Encoders used for UCF-101 are ImageNet pre-trained. Encoders of other datasets are trained from scratch. During the training, we use SGD with momentum ($0.9$) and set the learning rate at $1e-3$. A subset with $20\%$ training samples is randomly split in modality-level method. During modality valuation, for input modality set $C$, input of modalities not in $C$ are zeroed out, similar to related work~\cite{ghorbani2020neuron}. During testing, all modalities are taken as the model input. Detailed experimental settings, experiments about more than two modalities, and ablation studies about the subset scale, $f_s(\cdot)$ as well as $f_m(\cdot)$, are provided in \emph{Supp. Materials}.

\subsection{Comparison with multimodal fusion methods}

Here we first compare our methods with several representative multimodal fusion methods under deep frameworks: Concatenation~\cite{owens2018audio}, Summation, Decision fusion~\cite{simonyan2014two}, FiLM~\cite{perez2018film} and Gated~\cite{kiela2018efficient}. Besides, the early multimodal integration attempts, Bayesian network~\cite{blundell2015weight} and Multi-kernel Learning (MKL)~\cite{sikka2013multiple}, are also compared. To be fair, the uni-modal encoders of Bayesian network are ResNet-18 and features fed into MKL are extracted by pre-trained uni-modal encoders. Our sample-level and modality-level methods are based on Concatenation fusion in Table~\ref{tab:mmframeworks}.

\begin{figure*}[t]
\centering
    \begin{subfigure}[t]{.22\textwidth}
			\centering
			\includegraphics[width=\textwidth]{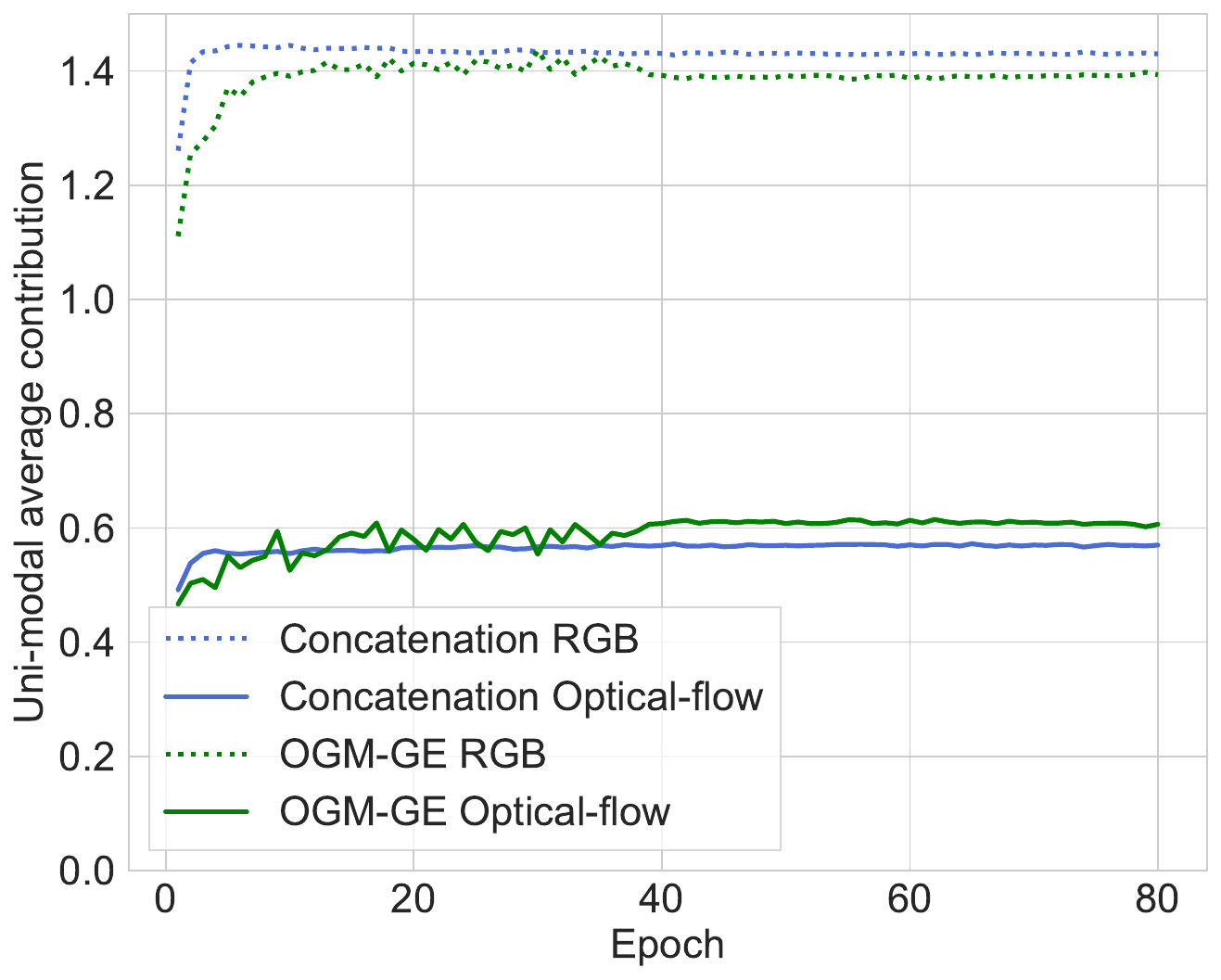}
			\caption{OGM-GE~\cite{peng2022balanced}.}
			\label{fig:ogm}
	\end{subfigure}
	\begin{subfigure}[t]{.22\textwidth}
			\centering
			\includegraphics[width=\textwidth]{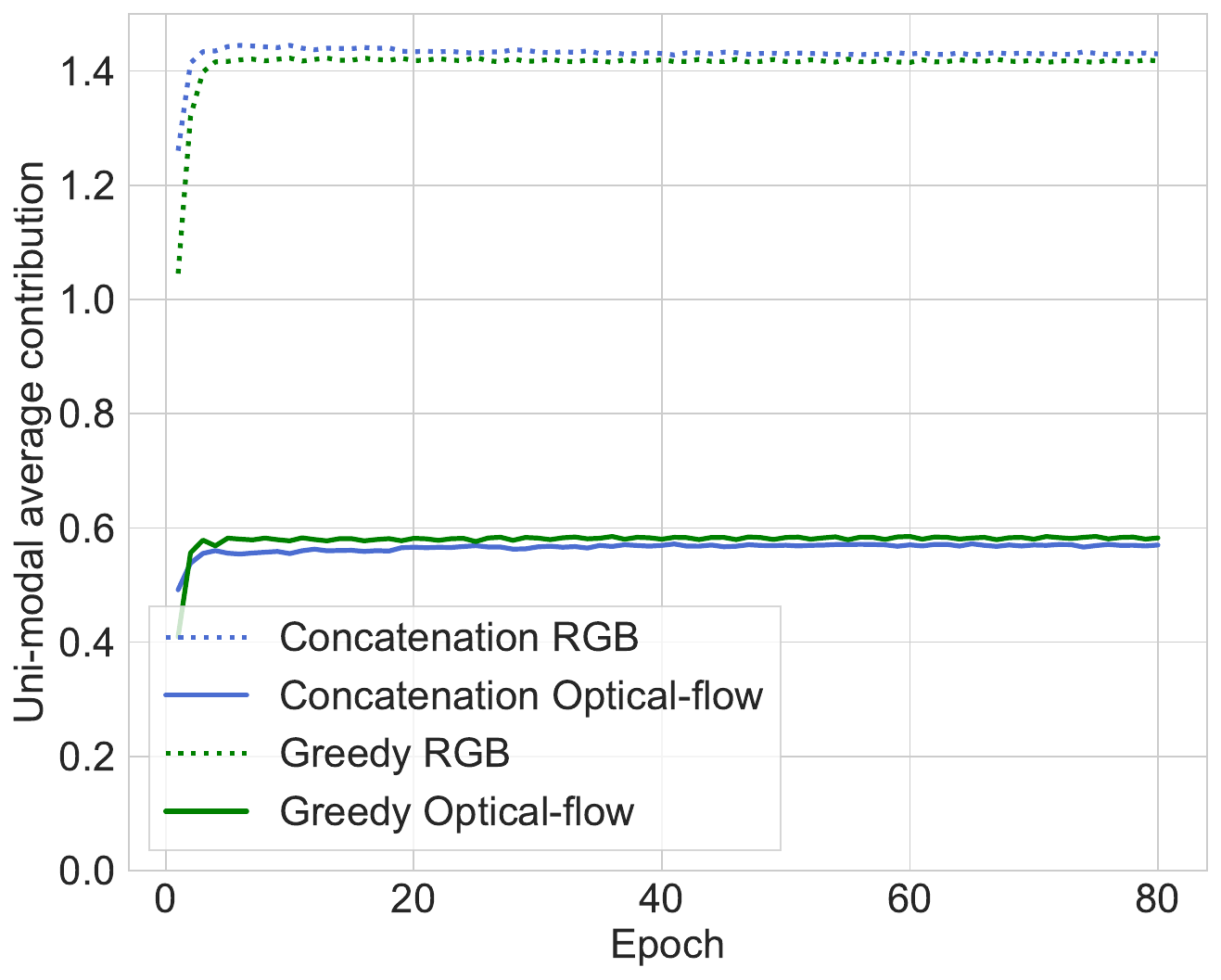}
			\caption{Greedy~\cite{wu2022characterizing}.}
			\label{fig:greedy}
	\end{subfigure}
 \begin{subfigure}[t]{.22\textwidth}
			\centering
			\includegraphics[width=\textwidth]{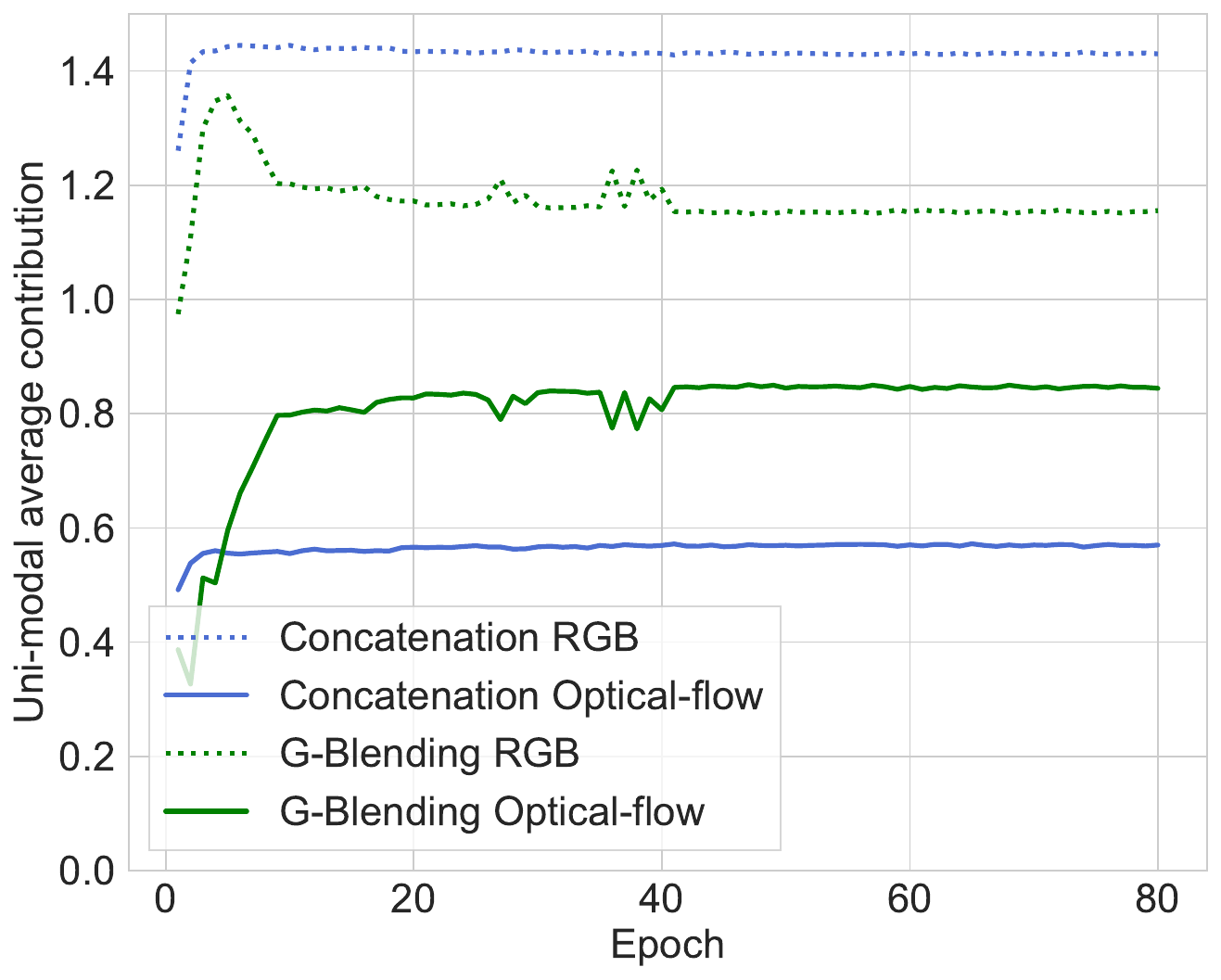}
			\caption{G-Blending~\cite{wang2020makes}.}
			\label{fig:g}
	\end{subfigure}
	\begin{subfigure}[t]{.22\textwidth}
			\centering
			\includegraphics[width=\textwidth]{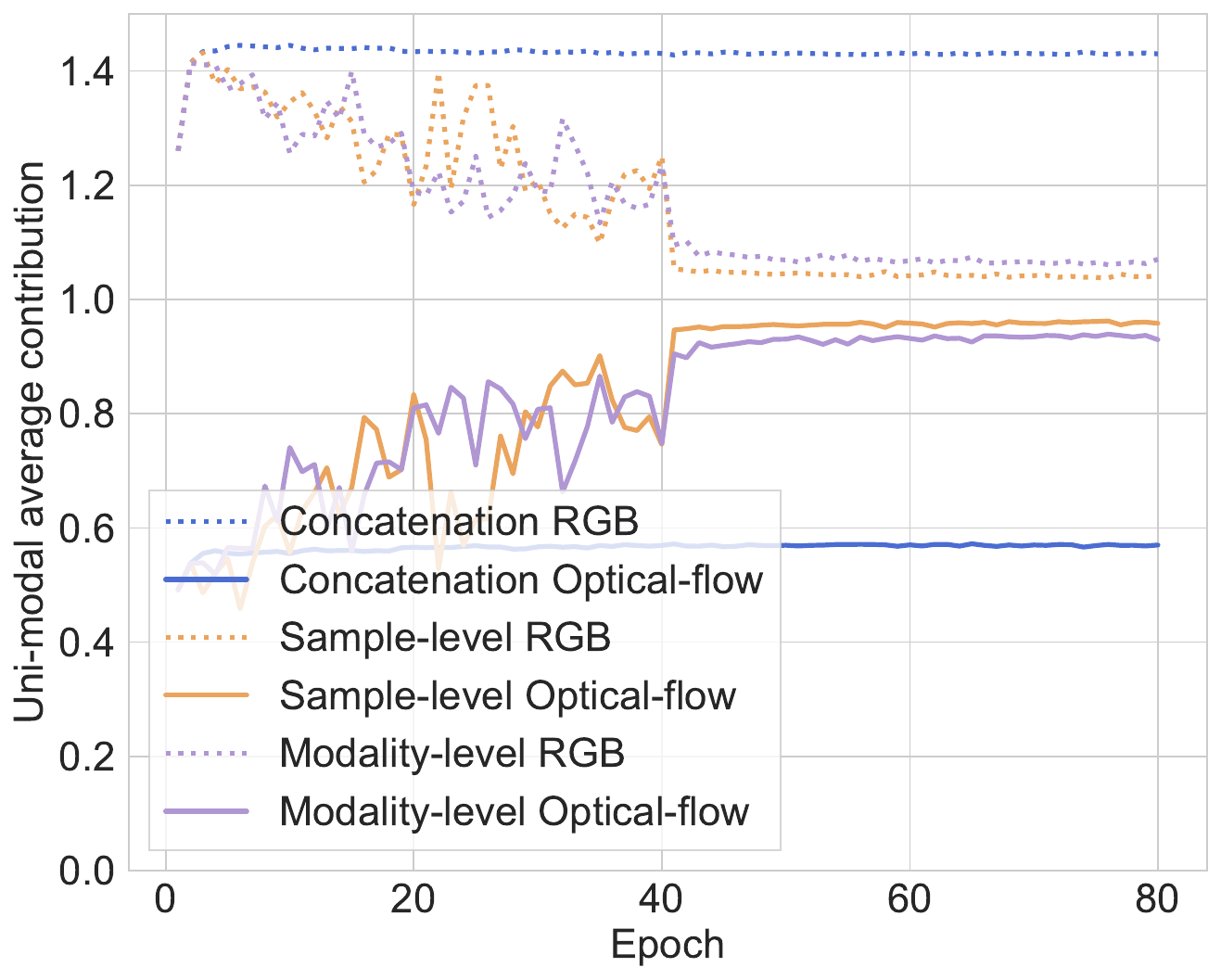}
			\caption{Ours.}
			\label{fig:m-our}
	\end{subfigure}
 \vspace{-0.5em}
    \caption{Average contribution of each modality over all training samples during training for OGM-GE, Greedy, G-Blending and our methods on the UCF-101 dataset.}
    \vspace{-0.5em}
    \label{fig:shap_modulation}
\end{figure*}

\begin{table}[t]
\centering
\begin{tabular}{c|cc|cc}
\bottomrule
\multirow{3}{*}{\textbf{Method}}  & \multicolumn{2}{c|}{\textbf{KS}}& \multicolumn{2}{c}{\textbf{UCF-101}} \\
&\multicolumn{2}{c|}{\textbf{(Audio+Visual)}}&\multicolumn{2}{c}{\textbf{(RGB+OF)}}\\
&\textbf{Acc}&\textbf{mAP}&\textbf{Acc}&\textbf{mAP}\\
\hline
Concatenation~ & 62.30 & 67.95 & 81.15 & 86.15 \\
Summation & 62.10 & 66.97  & 81.31  & 86.25 \\
Decision fusion & 62.65& 67.89 & 79.81 & 86.07 \\ 
FiLM~\cite{perez2018film} & 61.25 & 64.85 & 79.45  & 84.27  \\
Gated~\cite{kiela2018efficient} & 62.72 & 68.28 & 81.34 & 86.35 \\
\hline

Bayesian DNN~\cite{blundell2015weight} &60.79& 64.98 &80.04 & 84.95\\
Deep MKL*~\cite{sikka2013multiple} & 63.61& 69.69 & 82.64 & 87.96 \\
\hline
Sample-level & \textbf{66.92}& \underline{71.84} & \textbf{83.52} &\textbf{88.89}\\
Modality-level & \underline{66.65}& \textbf{72.68} & \underline{83.46} & \underline{88.75}  \\
 \toprule
\end{tabular}
\vspace{-1em}
\caption{\textbf{Comparison with different multimodal fusion methods.} Bold and underline represent the best and runner-up respectively. * denotes that the fed feature of Deep MKL model is extracted by pre-trained uni-modal encoders. Bayesian DNN is trained from scratch. OF denotes for optical flow. \emph{Due to limited space, experiments about more modalities, e.g., text modality, are provided in Supp. Materials.}}
\vspace{-1em}
\label{tab:mmframeworks}
\end{table}

Based on Table~\ref{tab:mmframeworks}, several observations can be revealed. Firstly, early multimodal integration methods are able to disclose their effectiveness after being equipped with extracted deep features, especially MKL, which even outperforms the Concatenation model. However, this improvement has a reliance on the quality of input features, and these methods are still hard to directly process raw data in large-scale. Secondly, both our sample-level and modality-level strategies improve multimodal cooperation by means of fine-grained modality valuation, achieving better model performance. Moreover, the fine-grained sample-level method tends to be superior. In contrast, the modality-level method is more efficient and sometimes even can be comparable to sample-level one.

\subsection{Comparison with imbalanced multimodal learning methods}
Recent studies have found that multimodal models often cannot jointly utilize all modalities well, and some imbalanced multimodal learning methods are proposed. They often control uni-modal optimization by estimating the discrepancy of the training stage or performance between modalities. Here we compare with these imbalanced multimodal learning methods, OGM-GE~\cite{peng2022balanced}, G-Blending~\cite{wang2020makes}, Greedy~\cite{wu2022characterizing}, PMR~\cite{fan2023pmr} and AGM~\cite{li2023boosting}. 
Our sample-level and modality-level methods are based on Concatenation fusion.

\begin{table}[t]
\centering

\setlength{\tabcolsep}{1.5mm}{
\begin{tabular}{c|ccc}
\bottomrule
\textbf{Method} & \textbf{KS} & \textbf{UCF-101} & \textbf{MM-Debiased} \\ \hline
Concatenation                    & 62.30       & 81.15            & 83.22                \\ \hline
OGM-GE~\cite{peng2022balanced}                           & 63.30       & 81.54            & 83.08 ($\downarrow$)               \\
G-Blending~\cite{wang2020makes}                       & 66.24       & 83.09            & 84.17                \\
Greedy~\cite{wu2022characterizing}                           & 62.37       & 81.25            & 83.08 ($\downarrow$)               \\
PMR~\cite{fan2023pmr}                              & 63.23       & 81.04 ($\downarrow$)            & 82.88 ($\downarrow$)               \\
AGM~\cite{li2023boosting}                              & 63.96       & 81.65            & 83.22 ($-$)               \\ \hline
Sample-level                     & \textbf{66.92}       & \textbf{83.52}            & \textbf{84.71}                \\
Modality-level                   & \underline{66.65}       & \underline{83.46}            & \underline{84.31}                \\ \toprule
\end{tabular}}
\vspace{-1em}
\caption{Accuracy of imbalanced multimodal learning methods, where bold and underline represent the best and runner-up respectively. $\downarrow$ indicates a performance drop compared with Concatenation baseline.}
\vspace{-1em}
\label{tab:modulation}
\end{table}

\noindent \textbf{Common case.} In many dataset, like Kinetics Sounds and UCF-101, as the former analysis, the low-contributing phenomenon has a dataset-level preference. Firstly, based on the results in Table~\ref{tab:modulation}, in this case with dataset-level modality preference, these methods often have a performance improvement. Among them, our methods outperform these imbalanced multimodal learning approaches. Although G-Blending~\cite{wang2020makes} achieves considerable performance, it needs to train an additional uni-modal classifier as the basis of modulation. Concretely, in fact, FLOPs of our methods reduce $1/4$ (sample-level method), even $1/2$ (modality-level method), compared to G-Blending.
Secondly, since our modality valuation is not limited to specific methods, the uni-modal contribution of other methods can also be observed. As Figure~\ref{fig:shap_modulation}, our methods exhibit superior mitigation of imbalanced uni-modal contributions, thereby highlighting our efficacy beyond mere final prediction.

\noindent \textbf{More balanced case.} However, under realistic scenarios, the modality discrepancy could vary across different samples, as samples shown in Figure~\ref{fig:teaser_audio} and~\ref{fig:teaser_visual}. To comprehensively evaluate imbalanced multimodal learning methods, we conduct experiments under the more balanced case, where dataset-level discrepancy is not apparent, but there are still sample-level modality discrepancy. We build the MM-Debiased dataset where the dataset-level discrepancy is no longer significant. As Figure~\ref{fig:shap-dataset}, the average contribution of each modality on the MM-Debiased dataset is more balanced than that on other curated dataset. Based on the results shown in Table~\ref{tab:modulation}, most imbalanced multimodal learning methods including OGM, Greedy and PMR are even worse than Concatenation baseline, since existing methods only consider the dataset-level modality preference. They could not capture the sample-level modality discrepancy. In contrast, our methods, especially our sample-level method, achieves considerable improvements. Our method can reasonably valuate fine-grained modality contribution, and targetedly enhance the learning of low-contributing modality.

\subsection{Comparison of sample-level modality valuation}
Beyond the model performance, we further conduct experiments to compare with existing imbalanced multimodal learning methods about sample-level modality valuation. In these methods, to modulate the uni-modal optimization, they also evaluate specific uni-modal properties. For example, G-Blending~\cite{wang2020makes} and Greedy~\cite{wu2022characterizing} inspect the uni-modal training process. AGM~\cite{li2023boosting} proposes to evaluate the modality contribution which is then used to modulate gradient. But they could not be used to evaluate the modal preference at the sample-level. The uni-modal confidence score used by OGM-GE~\cite{peng2022balanced} could be used to evaluate uni-modal performance at sample-level. However, this empirically designed score is hard to handle in realistic scenarios, like dominant modality could differ among samples within the same category, since its calculation could suffer from the dataset-level discrepancy, resulting in inaccurate results.

\begin{figure}[t!]
\centering 
	\begin{subfigure}[t]{.23\textwidth}
			\centering
			\includegraphics[width=\textwidth]{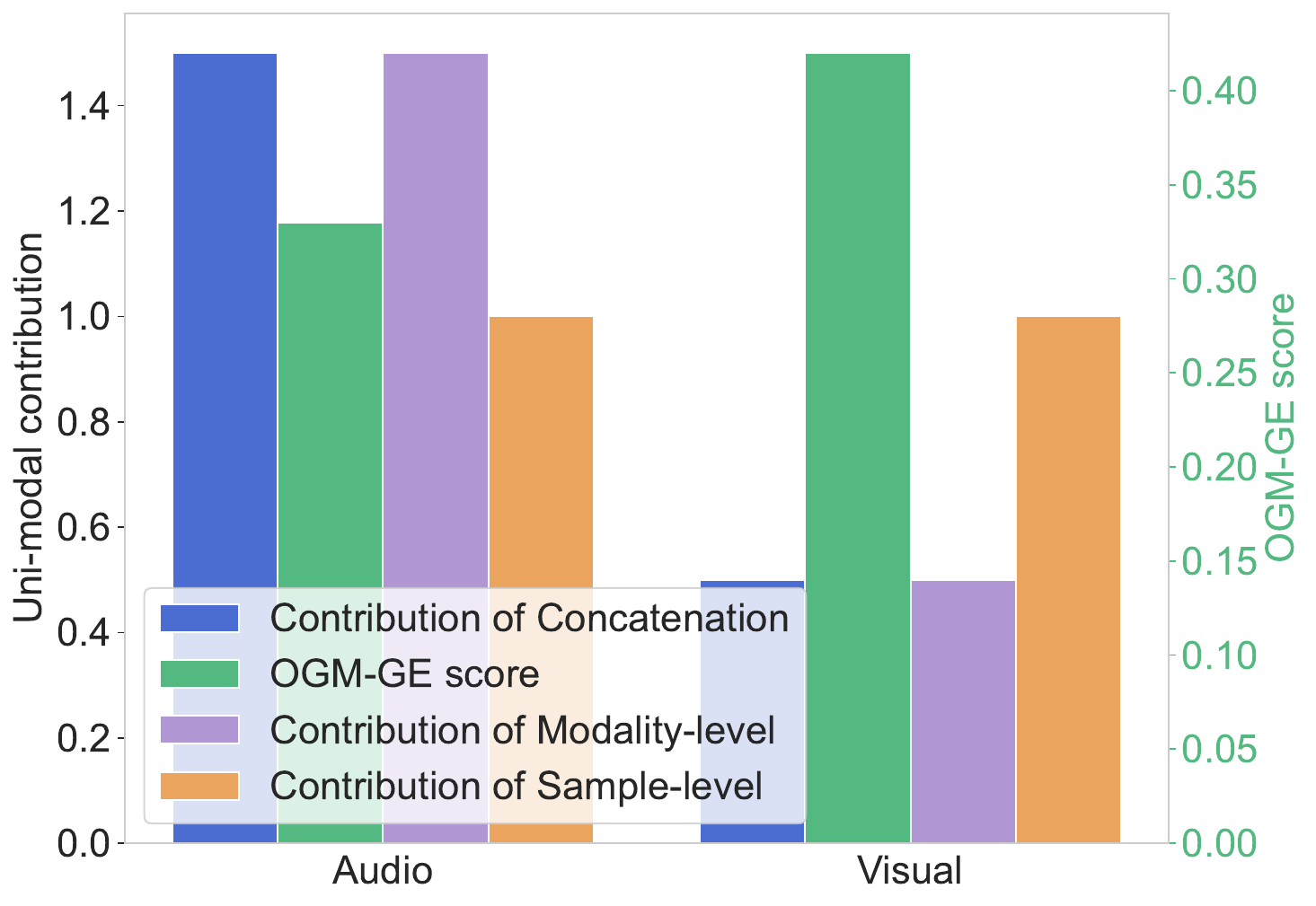}
			\caption{Valuation of \emph{Sample 1}.}
			\label{fig:case-1-shap}
	\end{subfigure}
	\begin{subfigure}[t]{.23\textwidth}
			\centering
			\includegraphics[width=\textwidth]{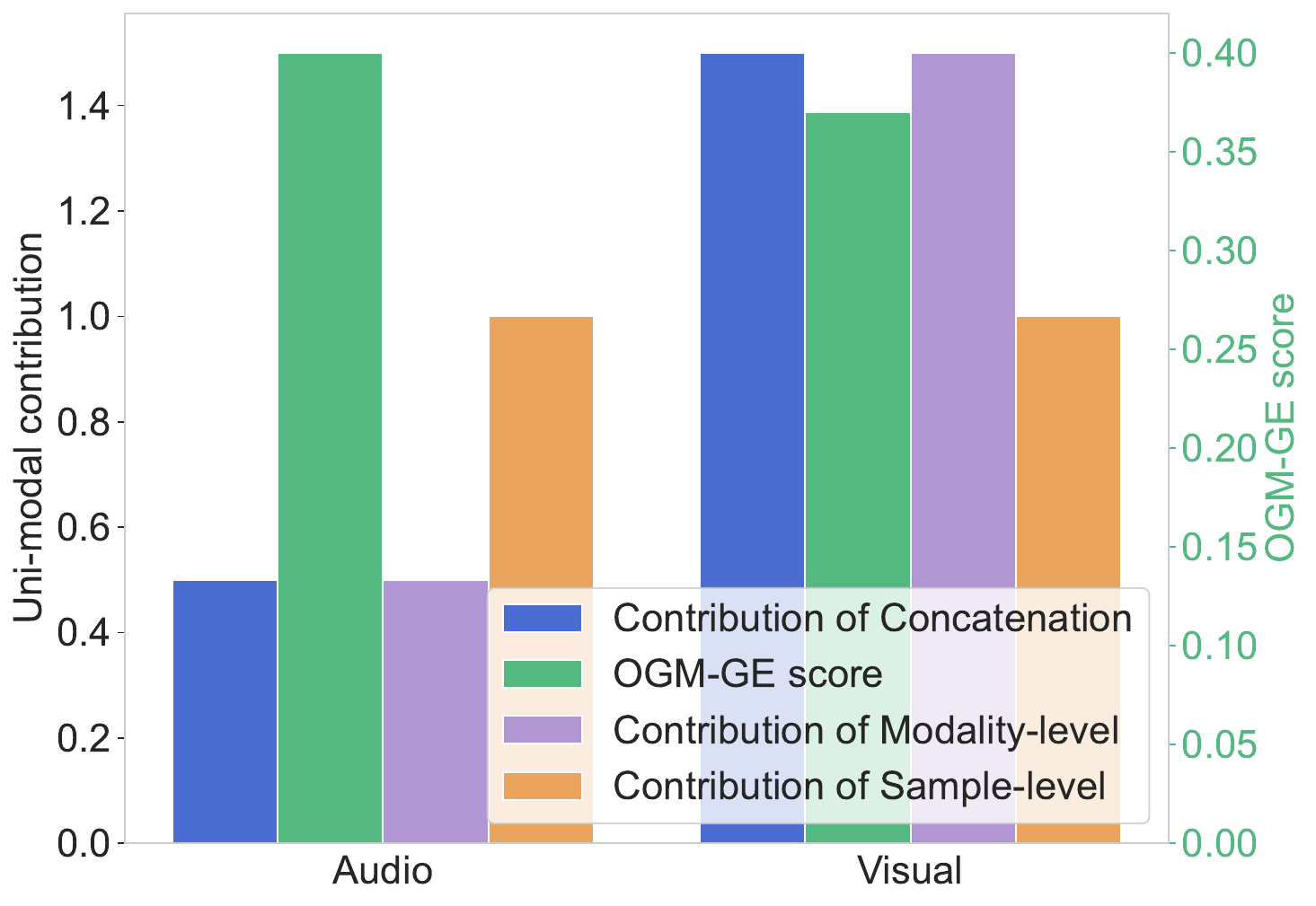}
			\caption{Valuation of \emph{Sample 2}.}
			\label{fig:case-2-shap}
	\end{subfigure}
     \vspace{-0.5em}
    \caption{Valuation of two samples of \emph{motorcycling} category.}
    \vspace{-1em}
    \label{fig:case}
\end{figure}

For example, for the two audio-visual samples of \emph{motorcycling} category in Figure~\ref{fig:teaser_audio} and~\ref{fig:teaser_visual}, the motorcycle in \emph{Sample 1} appears hard to be observed, while the wheel of motorcycle in \emph{Sample 2} appears clearly. These two samples receptively rely on audio or visual modality. Here we propose the modality valuation of different methods of these two samples. Results are shown in Figure~\ref{fig:case}. Based on the contribution of Concatenation baseline produced by our methods (blue bar in Figure~\ref{fig:case}), our valuation correctly reflects that these two samples rely on audio and visual signals, respectively. However, OGM-GE score provides the wrong results, assigning more confidence for the less informative visual signal of \emph{Sample 1} (green bar in Figure~\ref{fig:case-1-shap}). 

Moreover, our fine-grained sample-level method captures and accordingly adjusts the uni-modal learning, balancing this fine-grained modality discrepancy (orange bar in~Figure~\ref{fig:case}). Although our modality-level method fails to ease this discrepancy (purple bar in~Figure~\ref{fig:case}), it has an advantage in efficiency. These experiments also indicate that the sample-level and modality-level methods have their own advantages and applicable scenarios.

\begin{table}[t]
\centering
\setlength{\tabcolsep}{1mm}{
\begin{tabular}{c|ccc}
\bottomrule
  
\textbf{Method} & \textbf{KS} & \textbf{UCF-101} & \textbf{MM-Debiased} \\

\hline
Concatenation        & 62.30   &   81.15    &   83.22      \\
Concat-Sample       &  \textbf{66.92}  &   \textbf{84.71}  & \textbf{84.31}  \\
Concat-Modality     &  66.65  &   83.46   &  84.04 \\ \hline
CentralNet~\cite{vielzeuf2018centralnet}          & 67.35  &   83.97   & 85.23 \\
CentralNet-Sample   &  67.89  & \textbf{84.07} &  \textbf{85.39} \\
CentralNet-Modality &  \textbf{68.31}  & 84.05  & 85.26  \\ \hline
MMTM~\cite{joze2020mmtm}                & 64.23  &  80.67  &    84.31     \\
MMTM-Sample         &  \textbf{64.40}  &     \textbf{81.30}   &    \textbf{85.33}     \\
MMTM-Modality       &  64.34  &   81.23   &    84.71 \\ \hline
MBT~\cite{nagrani2021attention}        & 47.02   &   -   &    68.01     \\
MBT-Sample       & \textbf{47.53}  &  -   &  \textbf{68.70} \\
MBT-Modality     & 47.36   &   -   &  68.34 \\ \toprule
\end{tabular}}
\vspace{-0.5em}
\caption{\textbf{Accuracy of multimodal frameworks with cross-modal interaction modules.} Results of MBT on UCF-101 dataset could not be obtained since samples of it is hard to be trained from scratch but lacking suitable pre-trained transformer backbone.}
\label{tab:interaction}
\end{table}

\begin{figure*}[t]
\centering
    \begin{subfigure}[t]{.23\textwidth}
			\centering
			\includegraphics[width=\textwidth]{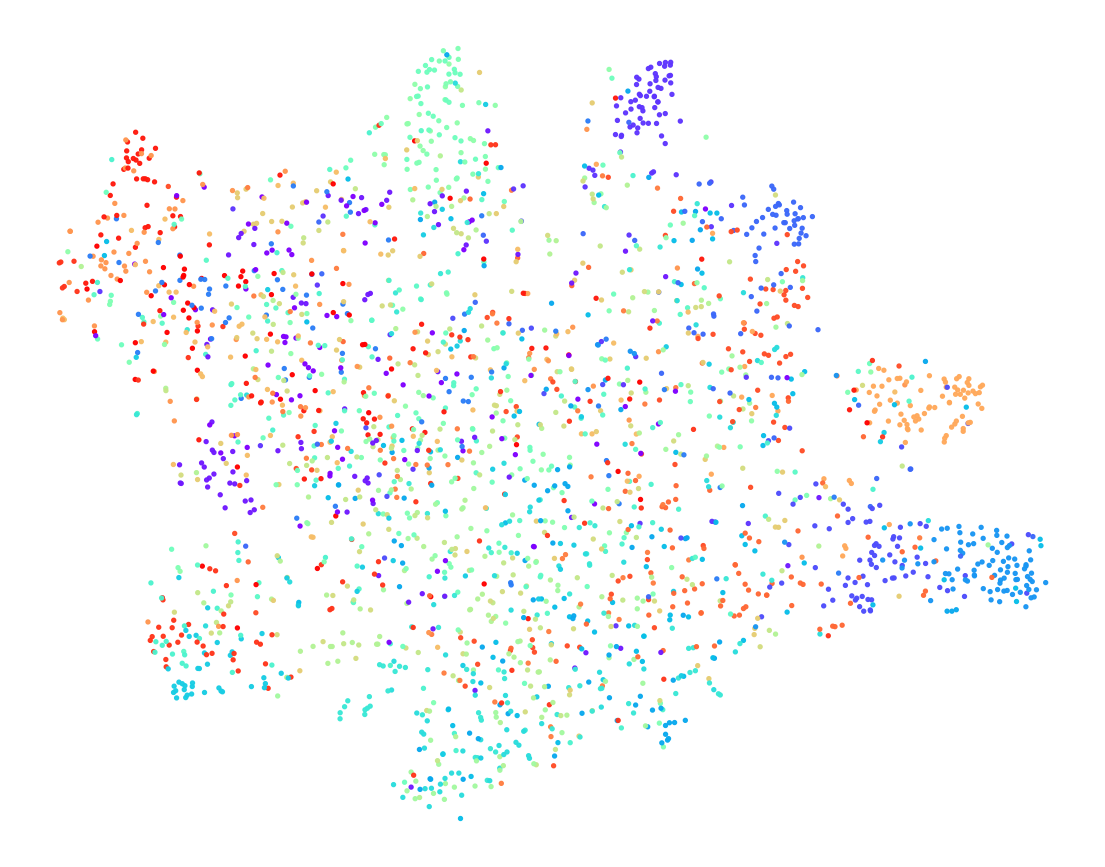}
			\caption{Concatenation.}
			\label{fig:tsne-baseline}
	\end{subfigure}
	    \begin{subfigure}[t]{.23\textwidth}
			\centering
			\includegraphics[width=\textwidth]{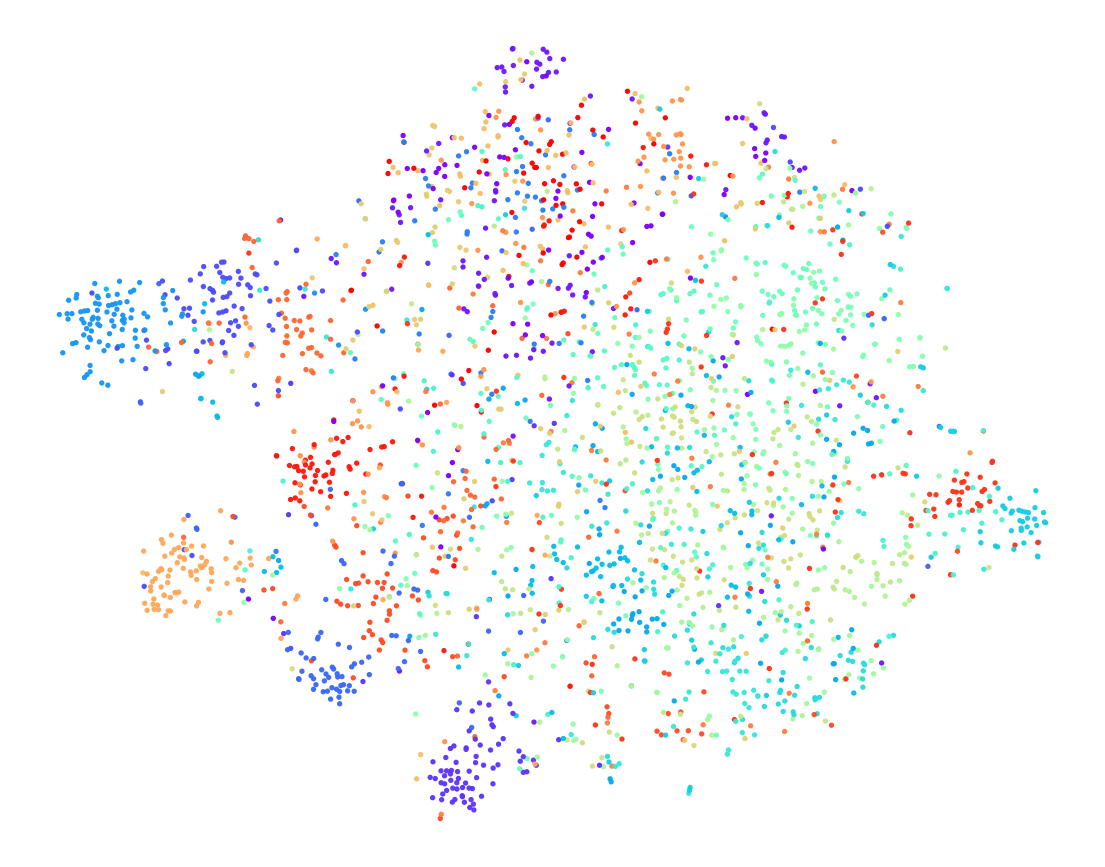}
			\caption{MMTM~\cite{joze2020mmtm}.}
			\label{fig:tsne-mmtm}
	\end{subfigure}
	\begin{subfigure}[t]{.23\textwidth}
			\centering
			\includegraphics[width=\textwidth]{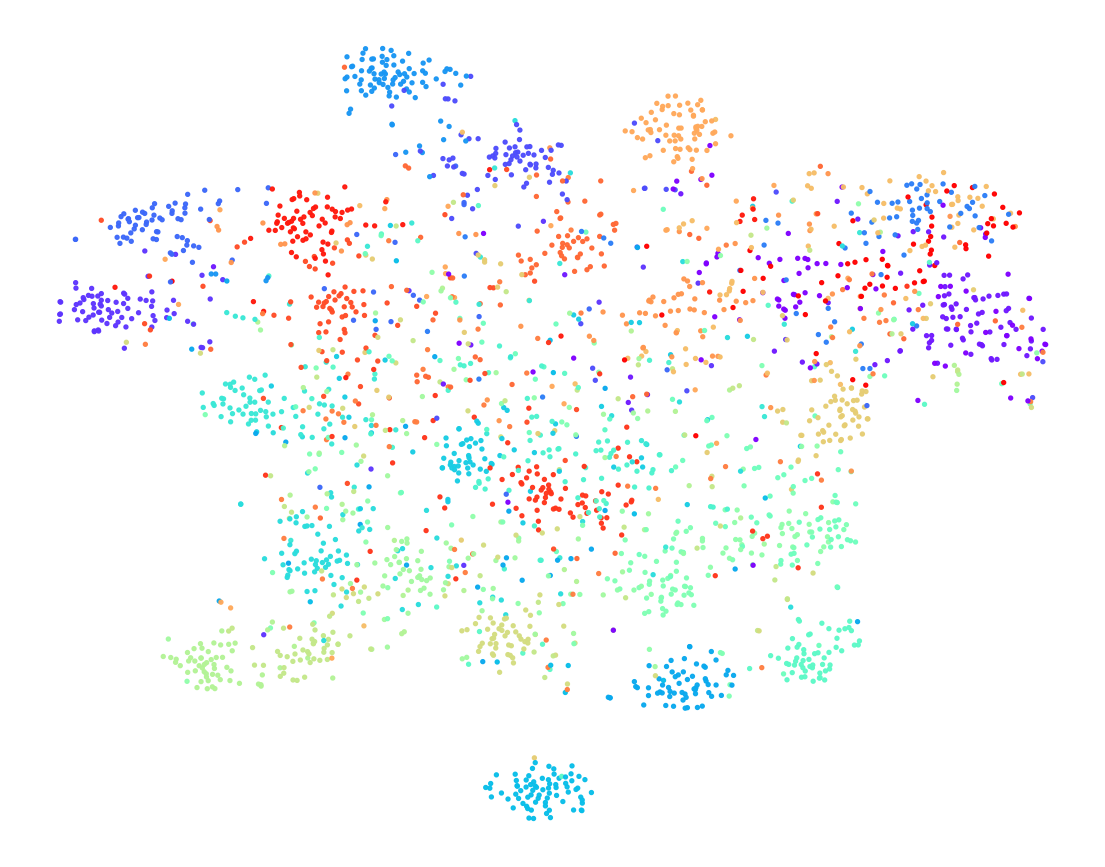}
			\caption{MMTM-Sample.}
			\label{fig:tsne-sample}
	\end{subfigure}
	\begin{subfigure}[t]{.23\textwidth}
			\centering
			\includegraphics[width=\textwidth]{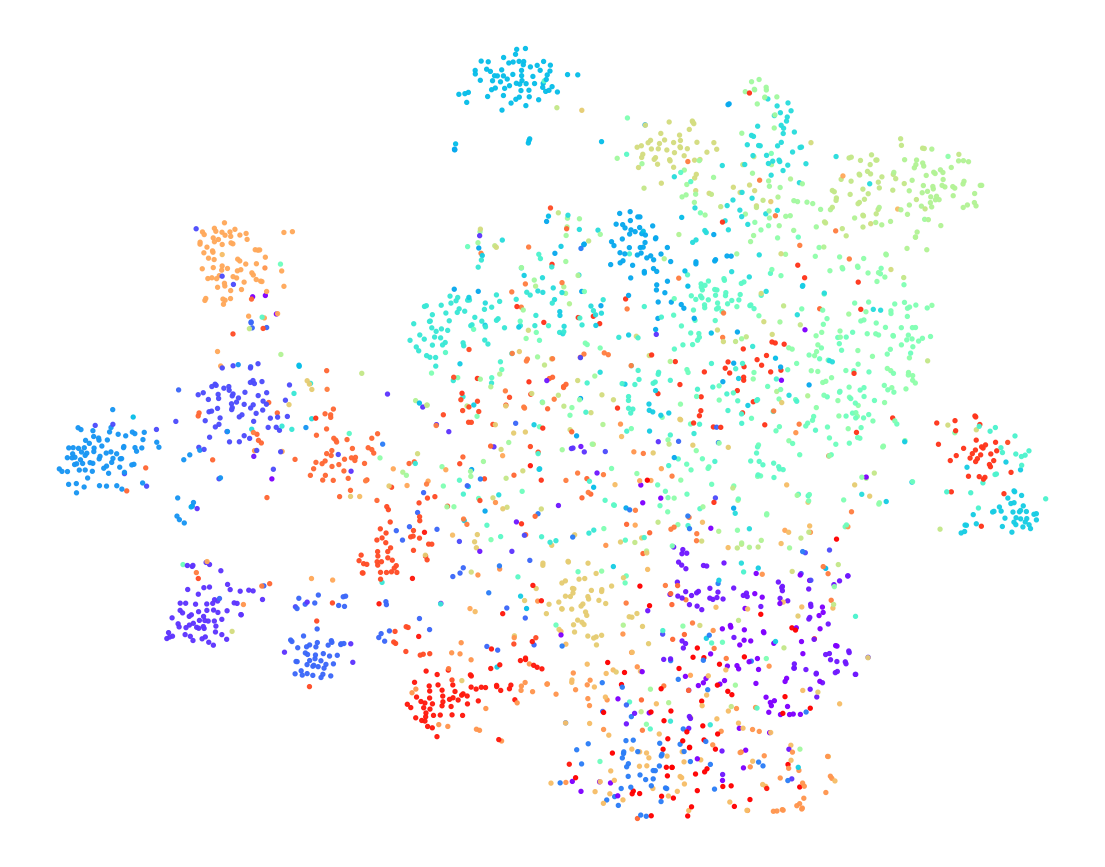}
			\caption{MMTM-Modality.}
			\label{fig:tsne-modality}
	\end{subfigure}
 \vspace{-0.5em}
    \caption{Visual feature distribution of Concatenation, MMTM, MMTM-Sample and MMTM-Modality, visualized by t-SNE~\cite{van2008visualizing} on Kinetics Sounds dataset. As Figure~\ref{fig:shap-dataset}, visual modality tends to be the low-contributing one. Categories are indicated in different colors.}
        \vspace{-1em}
    \label{fig:tsne}
\end{figure*}

\subsection{Complex cross-modal interaction scenarios}
As stated before, different from most existing imbalanced multimodal learning methods, our methods are not limited to simple fusion strategies. Here we first combine our sample-level and modality-level methods with two intermediate fusion methods, CentralNet~\cite{vielzeuf2018centralnet} and MMTM~\cite{joze2020mmtm}, to evaluate their effectiveness under these cross-modal interaction scenarios. Based on the results in Table~\ref{tab:interaction}, these cross-modal interaction modules indeed improve the model performance, compared with Concatenation baseline. This phenomenon demonstrates that the cross-modal interaction could implicitly deepen the cooperation between modalities by helping one modality make adjustments according to the feedback from others. 

In addition, our methods can be well applied to these more complex scenarios with cross-modal interaction, bringing performance improvement. One additional observation is that although the architecture limits model performance, our methods applied to the simple Concatenation fusion method could even have comparable results with more complex model designs, indicating it is simple-yet-effective. Furthermore, to qualitatively observe the quality of uni-modal representations, we visualize the feature distribution of overall low-contributing modality encoder on the Kinetics Sounds dataset (\emph{i.e.,} the visual modality). Results in Figure~\ref{fig:tsne} illustrate that the feature distribution is more discriminative in terms of action categories with cross-modal interaction, and this discriminative distribution can go a step further after being equipped with our methods.

Besides these modules based on CNN backbone, transformer networks also have cross-modal interaction. Here we combine our methods with the representative transformer model, MBT~\cite{nagrani2021attention}, to explore their effectiveness. Results are in Table~\ref{tab:interaction}. Models are trained from scratch. It should be noted that the performance of MBT is inferior to Concatenation with CNN backbone, since the transformer network is generally data-hungry, limiting its performance on these datasets without large enough samples, when being trained from scratch. But both our sample-level and modality-level methods can combine with and further enhance the performance of MBT. Overall, our methods could be well equipped with different cross-modal interaction modules, bringing performance enhancement.

\begin{table}[t]
\centering
\tabcolsep=0.14cm
\begin{tabular}{c|cc|cc}
\bottomrule
\multirow{2}{*}{\textbf{Method}}  & \multicolumn{2}{c|}{\textbf{KS}}& \multicolumn{2}{c}{\textbf{UCF-101}} \\
&\textbf{Acc}&\textbf{mAP}&\textbf{Acc}&\textbf{mAP}\\
\hline
Concatenation~ & 62.30 & 67.95 & 81.15 & 86.15  \\
\hline
Na\"ive re-sample & 64.69& 70.87 & 82.56 & 88.02 \\
Reversed re-sample & 59.03& 63.14 & 80.85 & 85.06\\
\hline
Sample-level & \textbf{66.92}& \underline{71.84} & \textbf{83.52} &\textbf{88.89}\\
Modality-level & \underline{66.65}& \textbf{72.68} & \underline{83.46} & \underline{88.75}  \\
\toprule
\end{tabular}
\caption{\textbf{Comparison with other re-sample strategies.} Bold and underline represent the best and runner-up respectively.}
\label{tab:resample}
\end{table}

\begin{figure}[t!]
\centering
    \begin{subfigure}[t]{.23\textwidth}
			\centering
			\includegraphics[width=\textwidth]{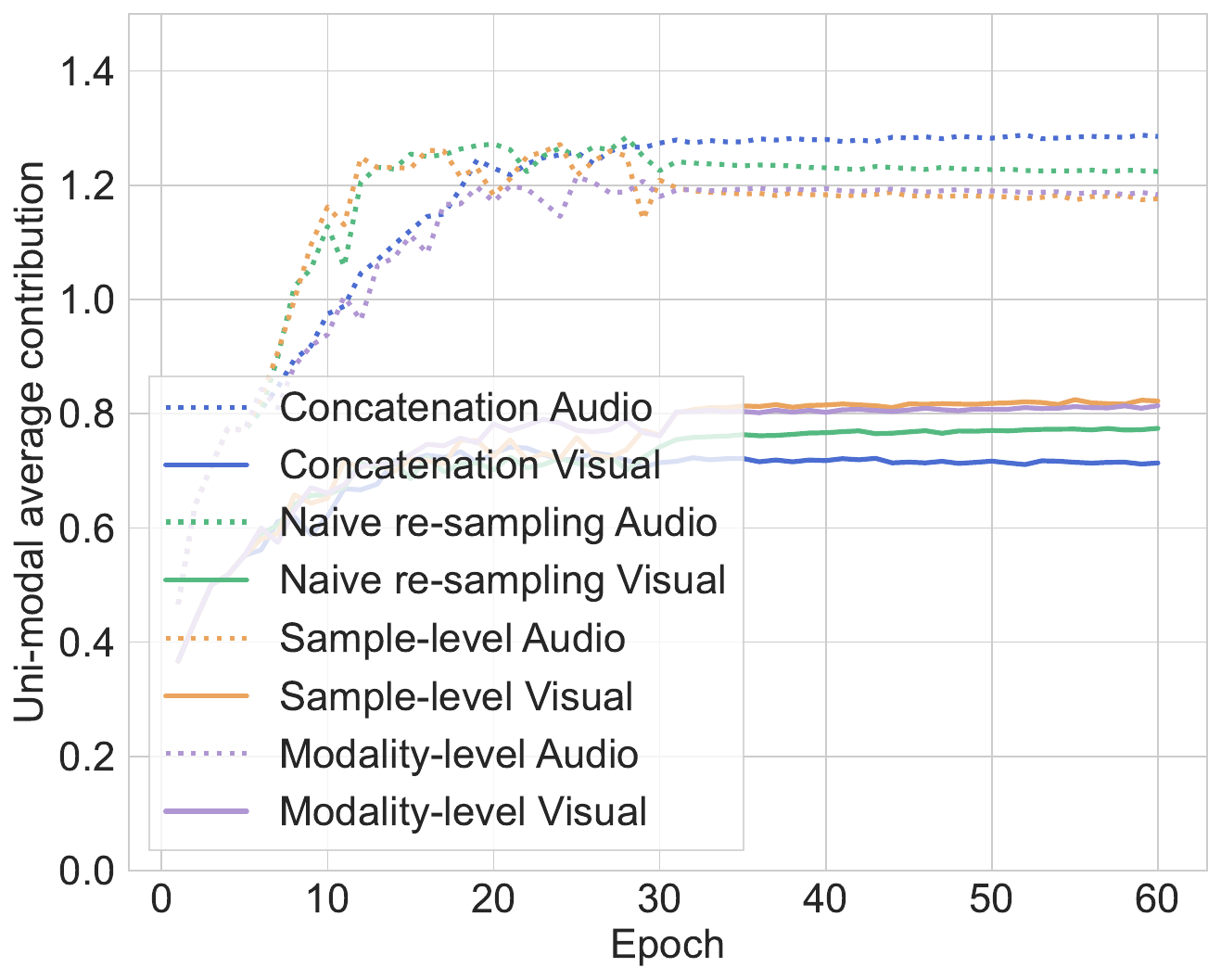}
			\caption{Na\"ive comparison.}
			\label{fig:naive}
	\end{subfigure}
	\begin{subfigure}[t]{.23\textwidth}
			\centering
			\includegraphics[width=\textwidth]{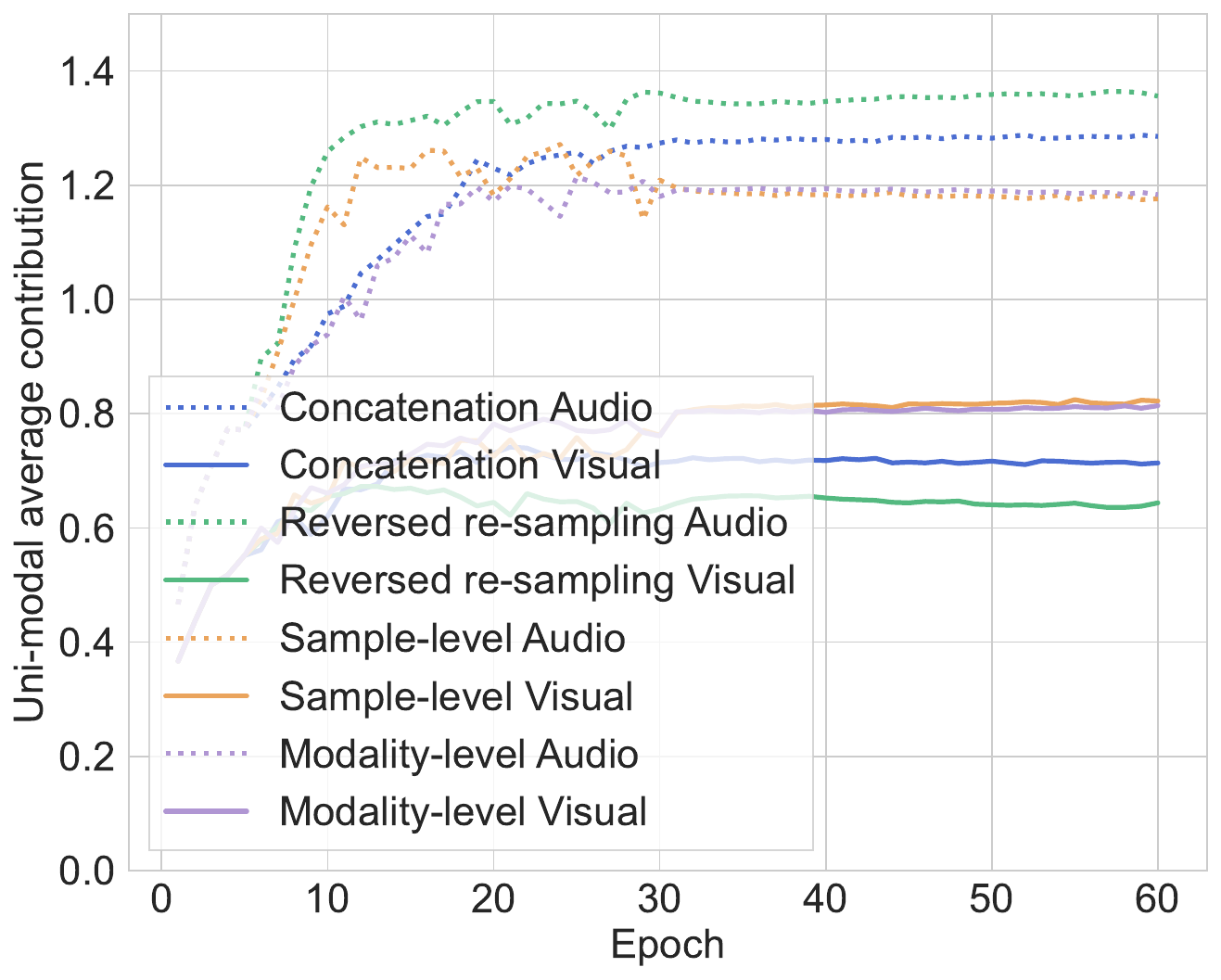}
			\caption{Reversed comparison.}
			\label{fig:reversed}
	\end{subfigure}
 \vspace{-0.5em}
    \caption{Average contribution of each modality over all training samples during training for Na\"ive re-sample and Reversed re-sample methods on the Kinetics Sounds dataset.}
    \vspace{-0.5em}
    \label{fig:final}
\end{figure}

\subsection{Comparison with other re-sample strategies}
To further validate our methods under the guidance of sample-level modality valuation, we compare with two related re-sample settings, na\"ive re-sample and reversed re-sample. Na\"ive re-sample is to randomly re-sample input of each modality with the same frequency as ours, while reversed re-sample is contrary to our methods, only re-sampling the data of modality with higher contribution. 

Based on Table~\ref{tab:resample}, na\"ive re-sample method can also increase model ability. The reason could be that this random uni-modal re-sample setting also potentially provides the chance for each modality to be trained individually, improving the discriminative ability of low-contributing modality. Hence, its contribution is accordingly boosted based on Remark~\ref{coro:enhance}. As the average uni-modal contribution shown in Figure~\ref{fig:naive}, na\"ive re-sample method indeed alleviates the low-contributing issues to some extent (the green line). Beyond that, our methods with targeted re-sample design under the guidance of sample-level modality valuation during training, take one step further (as the orange and purple lines). In addition, the failure of reversed re-sample setting (the green line in Figure~\ref{fig:reversed}), which runs counter to our analysis, also validates that it is our modality valuation guidance that matters, rather than the re-sample strategy itself.

\section{Discussion}
In this paper, we introduce a sample-level modality valuation metric to observe uni-modal contribution with the aid of theory in game theory. Two methods are proposed to recover the suppressed contribution of low-contributing modality, improving multimodal cooperation. Besides, there is also some further discussion.

\noindent \textbf{Natural difference between modalities.} In practice, there is a natural difference between modalities. For example, for an audio-visual sample \emph{drawing picture}, vision is naturally more discriminative than auditory. Hence, our methods could recover the suppressed contribution of low-contributing modality, but could not make the uni-modal contribution equal. Therefore, it is expected to evaluate and take this natural difference into consideration during improving multimodal cooperation in the future. 

\noindent \textbf{Imbalanced contribution in Multimodal Large Language Model.} Recently, the development of Multimodal Large Language Model has widely spread attention. However, in these models, like GPT-4V, there is also an imbalanced contribution issue. For example, results of GPT-4V are more likely to be misled by text modality~\cite{cui2023holistic}. To this end, the study about imbalanced uni-modal contribution is expected to extend to this case.

\noindent \textbf{Acknowledgements.} This research was supported by National Natural Science Foundation of China (NO.62106272).

{
    \small
    \bibliographystyle{ieeenat_fullname}
    \bibliography{main}
}


\end{document}